\documentclass[pdflatex,sn-mathphys-num]{sn-jnl}% Math and Physical Sciences Numbered Reference Style
\usepackage{graphicx}
\usepackage{multirow}%
\usepackage{amsmath,amssymb,amsfonts}%
\usepackage{amsthm}%
\usepackage{mathrsfs}%
\usepackage[title]{appendix}%
\usepackage{xcolor}%
\usepackage{textcomp}%
\usepackage{manyfoot}%
\usepackage{booktabs}%
\usepackage{algorithm}%
\usepackage{algorithmicx}%
\usepackage{algpseudocode}%
\usepackage{listings}%
\usepackage{adjustbox}
\usepackage{float}
\usepackage{tabularx}
\usepackage{subcaption}

\theoremstyle{thmstyleone}%
\theoremstyle{thmstyletwo}%

\theoremstyle{thmstylethree}%

\begin{document}

\title[Article Title]{C2L-Net: A Data-Driven Model for State-of-Charge Estimation of Lithium-Ion Batteries During Discharge}

\author*[1]{\fnm{Khoa} \sur{Tran}}\email{trandinhkhoa@tdtu.edu.vn}

\author[2]{\fnm{Tri} \sur{Le}}\email{tri.le@aiware.website}

\author[3]{\fnm{Nhu} \sur{Nguyen Gia}}\email{nguyengianhu@duytan.edu.vn}

\author[4,5]{\fnm{T.} \sur{Nguyen-Thoi}}\email{trung.nguyenthoi@vlu.edu.vn}

\author*[6,7]{\fnm{Vin} \sur{Nguyen-Thai}}\email{vin.nguyenthai@vlu.edu.vn}

\author[8,9]{\fnm{Duong} \sur{Tran Anh}}\email{duong.trananh@vlu.edu.vn}

\author[10]{\fnm{Hung-Cuong} \sur{Trinh}}\email{trinhhungcuong@tdtu.edu.vn}

\affil*[1]{\orgdiv{Data Science Laboratory, Faculty of Information Technology}, 
\orgname{Ton Duc Thang University}, 
\orgaddress{\city{Ho Chi Minh City}, \country{Vietnam}}}

\affil[2]{\orgdiv{AIWARE Limited Company}, 
\orgaddress{\city{Da Nang City}, \country{Vietnam}}}

\affil[3]{\orgdiv{School of Computer Science \& Artificial Intelligence}, 
\orgname{Duy Tan University}, 
\orgaddress{\city{Da Nang City}, \country{Vietnam}}}

\affil[4]{\orgdiv{Laboratory for Applied and Industrial Mathematics, Institute for Computational Science and Artificial Intelligence}, 
\orgname{Van Lang University}, 
\orgaddress{\city{Ho Chi Minh City}, \postcode{70000}, \country{Viet Nam}}}

\affil[5]{\orgdiv{Faculty of Mechanical, Electrical, and Computer Engineering, Van Lang School of Technology}, 
\orgname{Van Lang University}, 
\orgaddress{\city{Ho Chi Minh City}, \postcode{70000}, \country{Viet Nam}}}

\affil*[6]{\orgdiv{Laboratory for Computational Mechanics, Institute for Computational Science and Artificial Intelligence}, 
\orgname{Van Lang University}, 
\orgaddress{\city{Ho Chi Minh City}, \postcode{70000}, \country{Viet Nam}}}

\affil*[7]{\orgdiv{Faculty of Civil Engineering, Van Lang School of Technology}, 
\orgname{Van Lang University}, 
\orgaddress{\city{Ho Chi Minh City}, \postcode{70000}, \country{Viet Nam}}}

\affil[8]{\orgdiv{Laboratory of Environmental Sciences and Climate Change, Institute for Computational Science and Artificial Intelligence}, 
\orgname{Van Lang University}, 
\orgaddress{\city{Ho Chi Minh City}, \country{Vietnam}}}

\affil[9]{\orgdiv{Faculty of Environment, Van Lang School of Technology}, 
\orgname{Van Lang University}, 
\orgaddress{\city{Ho Chi Minh City}, \country{Vietnam}}}

\affil[10]{\orgdiv{Natural Language Processing and Knowledge Discovery Research Group, Faculty of Information Technology}, 
\orgname{Ton Duc Thang University}, 
\orgaddress{\city{Ho Chi Minh City}, \country{Vietnam}}}

\abstract{
Accurate state-of-charge (SOC) estimation is critical for the safe and efficient operation of lithium-ion batteries in battery management systems (BMS). Although data-driven approaches can effectively capture nonlinear battery dynamics, many existing methods rely on long historical input sequences, resulting in high computational cost and introducing padding-induced positional bias at the beginning of drive cycles. To address these limitations, we propose C2L-Net, a novel context-to-latest data-driven framework for realistic online SOC estimation using only a short historical window (20 s). Unlike existing short-receptive-field or long-history models, the proposed framework explicitly separates contextual encoding from latest-measurement updating, enabling both efficient temporal modeling and rapid adaptation to dynamic battery states. The proposed model incorporates a chunk-based feature extraction mechanism that combines Theta Attention Pooling with a Fourier-based Seasonality Basis to capture local temporal patterns while reducing sequence length. A causal context encoder, integrating a gated recurrent unit (GRU) with Causal Cosine Attention, models temporal dependencies without information leakage. Furthermore, a latest-measurement decoder, inspired by recursive filtering, updates the contextual state using the most recent measurement, enhancing responsiveness to dynamic operating conditions. Extensive experiments on a public lithium-ion battery drive-cycle dataset under multiple fixed-temperature conditions demonstrate that the proposed method achieves state-of-the-art or competitive accuracy while significantly improving computational efficiency. In particular, C2L-Net achieves up to 60$\times$ faster inference and requires fewer parameters than recent data-driven baselines, while maintaining robust performance across unseen driving profiles. These results highlight the effectiveness of the proposed architecture for practical deployment in real-time and resource-constrained BMS applications.}




\maketitle

\section{Introduction}\label{Introduction}
Lithium-ion batteries (LIBs) have been widely used in various applications, such as electric vehicles, portable devices, and renewable energy storage systems. Their adoption has been driven by advantages including high energy density, rapid charging capability, low self-discharge rate, long cycle life, and relatively low environmental impact~\cite{li2025lightweight}. Moreover, with global trend toward transitioning from combustion-based energy to green energy to reduce greenhouse gas emissions~\cite{hu2024state}, this also has led to the widespread adoption of LIBs.

During the use of LIBs, a battery management system (BMS)~\cite{saranathan2025navigating, wang2025integrated} is required to ensure safe and reliable operation. The BMS performs several key functions, including remaining useful life prediction~\cite{ge2024structural, cai2024deep}, state of health estimation~\cite{yang2025physics, sun2024state}, and state of charge (SOC) estimation~\cite{subashini2026physics, wong2024balancing}. Among these functions, SOC estimation has attracted considerable attention because it helps prevent overcharging and overdischarging~\cite{li2025lightweight}. Overcharging and overdischarging can cause several safety and degradation issues, such as battery swelling, capacity fading, internal short circuits, and even explosion, which seriously threaten battery safety and reliability. In general, SOC estimation approaches are classified into three categories: (1) direct measurement methods, (2) model-based methods, and (3) data-driven methods. 

Direct measurement methods are the simplest SOC estimation approaches, as they estimate SOC using directly measurable battery signals, such as voltage, current, and impedance. For example, \cite{pop2005state} reviewed several direct measurement techniques based on battery variables, including terminal voltage, equilibrium voltage, open-circuit voltage/electromotive force, impedance, and voltage relaxation behavior. However, the accuracy of these methods is often limited because the measured variables are strongly affected by temperature, discharge rate, battery aging, and relaxation effects. \cite{westerhoff2016electrochemical} proposed an electrochemical impedance spectroscopy (EIS)-based SOC estimation method, in which the battery impedance spectra are directly measured and then fitted to a simplified equivalent circuit. The extracted circuit parameters, such as \(R_{0}\), \(R_{C}\), and \(C\), are then used to estimate SOC. In their study, the battery was charged and discharged in 10\% SOC steps, followed by EIS measurements at each step. The results show that this method can improve SOC estimation, especially in the middle SOC range. Nevertheless, its accuracy can still be affected by temperature, battery aging, and the selected frequency range. To overcome the limitations of direct measurement methods, model-based methods have been developed to incorporate battery dynamics and improve SOC estimation accuracy under various operating conditions.

Model-based methods include equivalent circuit model (ECM)-based methods, electrochemical model-based methods~\cite{he2022comparative}, Kalman filter-based methods~\cite{cui2022extended}, particle filter-based methods~\cite{chen2019particle}, and observer-based methods~\cite{tang2017observer}. These approaches incorporate battery dynamics, such as voltage response, current flow, internal resistance, polarization effects, diffusion behavior, and nonlinear electrochemical reactions, to improve SOC estimation accuracy under different operating conditions. For example, \cite{bage2025enhanced} proposed a model-based SOC estimation method that combines a second-order Thevenin equivalent circuit model (SOT-ECM), adaptive forgetting factor recursive least squares (FFRLS), and a moving-step unscented transformed dual extended Kalman filter (MUT-DEKF). This method captures fast and slow battery dynamics, adapts model parameters under different temperatures, and improves robustness against noise and operating-condition variations. However, its performance still depends on the accuracy of the equivalent circuit model and requires parameter calibration, which may increase implementation complexity. \cite{chen2025design} proposed a model-based SOC estimation method using an equivalent circuit battery model and a robust unknown input observer (UIO) with predefined convergence time. The main advantage of this method is that it improves robustness against model disturbances and uncertainties while allowing the convergence time to be specified in advance, leading to faster and more accurate SOC estimation than a conventional sliding mode observer. However, its performance still depends on the accuracy of the battery model and observer parameter design, which may further increase implementation complexity. Therefore, data-driven methods have been increasingly developed to overcome these limitations by learning complex nonlinear battery behaviors directly from data without requiring explicit battery modeling or extensive parameter identification.

Data-driven methods rely entirely on measured battery data and typically use neural network models to learn the nonlinear relationship between input signals, such as voltage, current, and temperature, and the target SOC. For example, \cite{li2025lightweight} proposed a lightweight data-driven SOC estimation framework that combines sequential dual Savitzky--Golay (SG) filters, Bayesian optimization and hyperband (BOHB), and neural network pruning. The dual-SG filters smooth voltage, current, and temperature signals before inputting them into convolutional neural network (CNN)~\cite{purwono2022understanding}, long short-term memory (LSTM)~\cite{graves2012long}, and gated recurrent unit (GRU)~\cite{dey2017gate}-based models, while BOHB reduces hyperparameter-search time and pruning reduces model size for embedded-device deployment. \cite{bao2024ttsnet} separates voltage, current, and temperature into independent branches, uses a temporal transformer to capture temporal dependencies, and applies attention-guided feature fusion plus Kalman filtering to improve SOC estimation accuracy. However, although the model is robust and accurate in experimental settings, it may be less realistic for real-world deployment because it still uses recurrent processing at each time step and relies on heavy feature extraction, which can increase computational cost. \cite{yao2025multi} proposes a multi-scale temporal convolutional network (MSTCN) for online SOC estimation. It uses several temporal convolutional network branches with different receptive fields to extract short-term and long-term battery features, and then applies cross-scale self-attention to dynamically fuse these features. However, the very long receptive field, 3276.8 seconds, may introduce implicit positional leakage because the input history is unavailable at the beginning of a drive cycle and must therefore be padded. Since the paper only evaluates discharge profiles that start at 100\% SOC and end at 0\% SOC, the use of a very long receptive field cause the beginning of each sequence to contain a large amount of zero padding. As a result, high-SOC samples are associated with long zero-padded histories, whereas low-SOC samples appear later in the sequence and contain little or no padding. Therefore, the model may partially learn the position within the discharge profile from the padding pattern, rather than learning SOC purely from real voltage, current, and temperature dynamics. The paper states that causal TCNs insert padding at the beginning of input sequences to align outputs with time steps, and it also uses receptive fields up to 3276.8 seconds, which supports this concern.

% ------------------------------------------------

Despite the significant progress achieved by existing SOC estimation methods, several critical limitations remain for practical deployment in real-world battery management systems. First, many recent data-driven approaches rely on long historical input sequences to capture temporal dependencies, which increases computational cost and memory usage, making them less suitable for real-time and resource-constrained environments. Moreover, the use of long input windows often requires zero padding at the beginning of drive cycles, introducing implicit positional bias. As a result, models may partially learn artificial patterns related to sequence position rather than true battery dynamics, which can degrade generalization performance in realistic operating conditions. Second, although advanced architectures such as transformers and multi-scale temporal models have improved estimation accuracy, they often involve complex feature extraction pipelines and high computational overhead. This complexity limits their applicability in embedded systems, where low latency and efficiency are essential. 

Therefore, there is a need for a lightweight and efficient SOC estimation
framework that (1) avoids dependence on long historical sequences and
(2) maintains high estimation accuracy under dynamic operating conditions
while enabling fast inference. To address these challenges, we propose
\textbf{C2L-Net}. The main contributions of this study are summarized as
follows:

\begin{itemize}
    \item \textbf{A SOC estimation framework:}
    We propose C2L-Net, a context-to-latest framework that accurately estimates SOC using only a \(20\,\mathrm{s}\) historical window. The input sequence is processed by a chunking module to reduce the computational cost of subsequent temporal modeling while maintaining high prediction accuracy.
    
    \item \textbf{A Feature Extraction Module:}
    We introduce a Feature Extraction Module comprising Theta Attention
    Pooling, a Fourier-based Seasonality Basis, and a Fusion module to capture
    informative temporal representations with low computational cost.

    \item \textbf{A Context Encoder:}
    We introduce a Context Encoder comprising a GRU and Causal Cosine
    Attention to model temporal dependencies and extract contextual
    information from the short historical window.

    \item \textbf{A Latest-Measurement Decoder:}
    We introduce a Latest-Measurement Decoder comprising a GRUCell and an MLP
    to update the SOC estimate using the latest measurement together with the
    context state obtained from the Context Encoder.

    \item \textbf{Comprehensive experimental validation:}
    Extensive experiments on a public drive-cycle dataset demonstrate the
    effectiveness of the proposed framework across multiple ambient
    temperatures and dynamic drive profiles.
\end{itemize}

The remainder of this paper is organized as follows. Section~\ref{Proposed_Method} presents the proposed method, including the overall architecture of the proposed C2L-Net and its detailed components. Section~\ref{Dataset} describes the dataset used in this study. Section~\ref{Experiments_Discussion} reports the experimental setup, evaluation metrics, baseline comparisons, and discussion of the obtained SOC estimation results, and also discusses potential future work. Finally, Section~\ref{Conclusion} concludes this paper.

\section{Proposed Method}\label{Proposed_Method}

\subsection{C2L-Net Overview}

\begin{figure}[H]
    \centering
    \includegraphics[width=\textwidth]{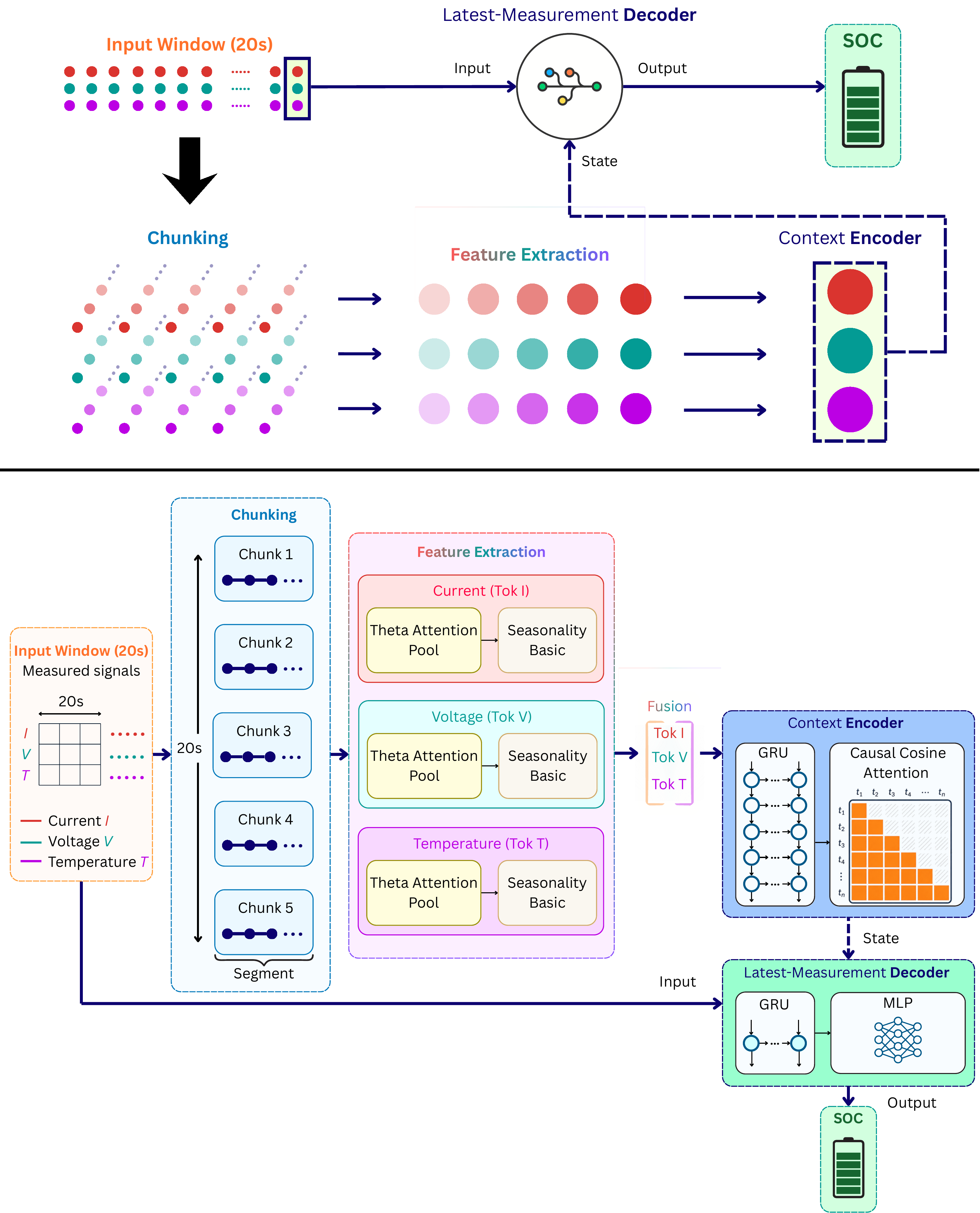}
    \caption{Overall architecture of the proposed C2L-Net.}
    \label{Overall_flow}
\end{figure}

At each time step \(t\), C2L-Net inputs a short sequence of current, voltage, and temperature measurements:
\begin{equation}
    \mathbf{X}_{t-L+1:t}
    =
    \left[
    \mathbf{x}_{t-L+1},
    \mathbf{x}_{t-L+2},
    \ldots,
    \mathbf{x}_{t}
    \right]
    \in \mathbb{R}^{L \times C},
\end{equation}
where \(L\) is the input window length (\(L = 200\), corresponding to 20\,s in the current setting) and \(C\) denotes the number of input variables, including current, voltage, and temperature (\(C=3\)). Each input vector is defined as
\(
    \mathbf{x}_{t} = [I_t, V_t, T_t],
\)
where \(I_t\), \(V_t\), and \(T_t\) represent the measured current, voltage, and temperature at time step \(t\), respectively. During online inference, the input window is updated in a sliding manner by one measurement step at a time, while maintaining a fixed-length recent window (e.g., 20\,s). 

The goal of the proposed C2L-Net model is to learn a nonlinear mapping function \(f_{\theta}(\cdot)\) to estimate the current SOC:
\begin{equation}
    \widehat{SOC}_{t}
    =
    f_{\theta}
    \left(
    \mathbf{X}_{t-L+1:t}
    \right).
\end{equation}
The overall pipeline of the model is illustrated in Fig.~\ref{Overall_flow}, which shows the overall process in the upper part and the detailed architecture in the lower part. The architecture consists of four main components: (1) chunking, (2) feature extraction, (3) context encoder, and (4) latest-measurement decoder. Specifically, the proposed model takes recent current, voltage, and temperature measurements as input and applies a chunking module to divide the input sequence into several segments. For each signal type, a Theta Attention Pool module extracts compact temporal representations, while a Seasonality Basis module generates a single representative value used as a token. These tokens are then processed by a context encoder composed of a GRU and Causal Cosine Attention, which captures relevant historical temporal dependencies and produces a contextual state. Finally, a latest-measurement decoder, consisting of a GRU cell and a multilayer perceptron (MLP), combines the latest measurement with the contextual state to generate the SOC estimate.

\subsection{Chunking}

% To capture local temporal patterns while keeping the input realistic for online estimation, the input sequence is divided into \(N\) chunks:
% \begin{equation}
%     \mathbf{X}_{t-L+1:t}
%     \rightarrow
%     \left\{
%     \mathbf{X}^{(1)},
%     \mathbf{X}^{(2)},
%     \ldots,
%     \mathbf{X}^{(N)}
%     \right\},
% \end{equation}
% where each chunk has length
% \begin{equation}
%     L_c = \frac{L}{N}.
% \end{equation}
The chunking module divides the input window \(\mathbf{X}_{t-L+1:t}\) into five chunks \((N = 5)\) to reduce the sequence length processed by the GRU in the context encoder. This design improves computational efficiency while still preserving sufficient information at each step, since processing  time-step sequences with recurrent models such as GRU and LSTM, or Transformer, is slow and less suitable for real-world BMS applications. This chunking module allows the model to extract local intra-temporal information from a segment. After chunking, the input sequence is transformed from
\begin{equation}
    \mathbf{X}_{t-L+1:t} \in \mathbb{R}^{L \times C}
\end{equation}
to
\begin{equation}
    \mathbf{X}_{\mathrm{chunk}}
    \in
    \mathbb{R}^{N \times L_c \times C},
\end{equation}
where \(L_c=L/N\).

\subsection{Feature Extraction}

After chunking, the feature extraction module takes the chunked representation
\(
    \mathbf{X}_{\mathrm{chunk}}
\)
as input, and extracts features from current, voltage, and temperature through three separate branches. This signal-wise design allows the model to learn the temporal characteristics of each measured variable independently before token fusion. The chunked input is first separated into three signal-specific tensors:
\begin{equation}
    \mathbf{X}_{\mathrm{chunk}}
    =
    \left[
    \mathbf{X}_{I},
    \mathbf{X}_{V},
    \mathbf{X}_{T}
    \right],
\end{equation}
where
\(
    \mathbf{X}_{I}, \mathbf{X}_{V}, \mathbf{X}_{T}
    \in
    \mathbb{R}^{N \times L_c}.
\)
Each row of these matrices corresponds to one chunk. Therefore, for the \(n\)-th chunk, the signal-specific sequences are defined as
\begin{equation}
    \mathbf{s}^{(n)}_{I} = \mathbf{X}_{I}^{(n,:)}, 
    \qquad
    \mathbf{s}^{(n)}_{V} = \mathbf{X}_{V}^{(n,:)}, 
    \qquad
    \mathbf{s}^{(n)}_{T} = \mathbf{X}_{T}^{(n,:)},
\end{equation}
where
\(
    \mathbf{s}^{(n)}_{I}, \mathbf{s}^{(n)}_{V}, \mathbf{s}^{(n)}_{T}
    \in \mathbb{R}^{L_c}.
\)
More generally, \(\mathbf{s}^{(n)} \in \{\mathbf{s}^{(n)}_{I}, \mathbf{s}^{(n)}_{V}, \mathbf{s}^{(n)}_{T}\}\) denotes the current, voltage, or temperature sequence of the \(n\)-th chunk. 

\paragraph{Theta Attention Pool}
The Theta Attention Pool module is to compress each chunk into a compact representation while preserving the most informative time steps inside the chunk. Instead of treating all time steps equally, the module learns attention weights to emphasize important local variations in current, voltage, or temperature. Specifically, each time step of \(\mathbf{s}^{(n)}\) is first projected into a hidden representation:
\begin{equation}
    \mathbf{h}^{(n)}_{\tau}
    =
    \mathbf{W}_{p}s^{(n)}_{\tau}
    +
    \mathbf{b}_{p},
    \qquad
    \tau = 1,2,\ldots,L_c,
\end{equation}
where \(\mathbf{h}^{(n)}_{\tau} \in \mathbb{R}^{d}\), and \(d\) is the hidden dimension. Therefore, the hidden representation of one chunk is
\(
    \mathbf{H}^{(n)}
    =
    \left[
    \mathbf{h}^{(n)}_{1},
    \mathbf{h}^{(n)}_{2},
    \ldots,
    \mathbf{h}^{(n)}_{L_c}
    \right]
    \in
    \mathbb{R}^{L_c \times d}.
\)
The attention score for each time step is then calculated as
\begin{equation}
    e^{(n)}_{\tau}
    =
    \mathbf{W}_{a}
    \mathbf{h}^{(n)}_{\tau}
    +
    b_a,
\end{equation}
where \(e^{(n)}_{\tau} \in \mathbb{R}\). The normalized attention weight is computed by
\begin{equation}
    \alpha^{(n)}_{\tau}
    =
    \frac{
    \exp
    \left(
    e^{(n)}_{\tau}
    \right)
    }
    {
    \sum_{j=1}^{L_c}
    \exp
    \left(
    e^{(n)}_{j}
    \right)
    }.
\end{equation}
The chunk-level feature representation is obtained using weighted pooling:
\begin{equation}
    \mathbf{z}^{(n)}
    =
    \sum_{\tau=1}^{L_c}
    \alpha^{(n)}_{\tau}
    \mathbf{h}^{(n)}_{\tau},
\end{equation}
where \(\mathbf{z}^{(n)} \in \mathbb{R}^{d}\). 

\paragraph{Seasonality Basic}
The Seasonality Basis module, inspired by \cite{oreshkin2019n}, maps the chunk-level feature representation into a coefficient vector, which is then used to generate a token that captures local periodic and trend-related temporal patterns. Specifically, the chunk-level feature representation is first mapped into a coefficient vector:
\begin{equation}
    \boldsymbol{\theta}^{(n)}
    =
    \phi
    \left(
    \mathbf{z}^{(n)}
    \right),
\end{equation}
where $\phi(\cdot)$ denotes a feed-forward network implemented by two linear layers with a ReLU activation. The coefficient vector has the shape
\(
    \boldsymbol{\theta}^{(n)}
    \in
    \mathbb{R}^{K_{\theta}},
    K_{\theta}=1+2K,
\)
where \(K\) is the number of harmonics (set to 10 in this work). To enhance the representation of local temporal variations, the coefficient vector is projected using a Fourier-based Seasonality Basis. To enhance the representation of local temporal variations, the coefficient vector is projected using a Fourier-based Seasonality Basis. The basis matrix is defined as
\begin{equation}
    \mathbf{B}
    =
    \left[
    \mathbf{1},
    \cos(2\pi t),
    \sin(2\pi t),
    \ldots,
    \cos(2\pi Kt),
    \sin(2\pi Kt)
    \right],
\end{equation}
where
\(
    \mathbf{B}
    \in
    \mathbb{R}^{K_{\theta} \times H}.
\)
Here, \(H\) denotes the token length produced by the Seasonality Basis. The token representation of each chunk is obtained by
\begin{equation}
    \mathbf{o}^{(n)}
    =
    \boldsymbol{\theta}^{(n)}
    \mathbf{B},
\end{equation}
where
\(
    \mathbf{o}^{(n)}
    \in
    \mathbb{R}^{H}.
\)
This operation is performed independently for current, voltage, and temperature:
\begin{equation}
    \mathbf{o}^{(n)}_{I}
    =
    \boldsymbol{\theta}^{(n)}_{I}
    \mathbf{B}_{I},
    \qquad
    \mathbf{o}^{(n)}_{V}
    =
    \boldsymbol{\theta}^{(n)}_{V}
    \mathbf{B}_{V},
    \qquad
    \mathbf{o}^{(n)}_{T}
    =
    \boldsymbol{\theta}^{(n)}_{T}
    \mathbf{B}_{T}.
\end{equation}
Each chunk produces a token $\mathbf{o}^{(n)} \in \mathbb{R}^{H}$. 
The token sequence is obtained by stacking all chunk-level tokens:
\begin{equation}
    Tok =
    \left[
    \mathbf{o}^{(1)}, \mathbf{o}^{(2)}, \ldots, \mathbf{o}^{(N)}
    \right]^\top
    \in \mathbb{R}^{N \times H}.
\end{equation}
After processing all \(N\) chunks, the signal-wise token sequences are obtained as
\begin{equation}
    Tok_I, Tok_V, Tok_T
    \in
    \mathbb{R}^{N \times H}.
\end{equation}
In this study, $H=1$, so each signal-wise token sequence has the shape
\(
    Tok_I, Tok_V, Tok_T \in \mathbb{R}^{5 \times 1}.
\)
This setting reduces token dimensionality and improves computational efficiency, with each token representing the compact state of a chunk.

\paragraph{Fusion}
To integrate information from different signal types into a unified representation, the current, voltage, and temperature tokens are stacked to form a compact token sequence:
\begin{equation}
    \mathbf{Z}
    =
    \text{Stack}
    \left(
    Tok_I,
    Tok_V,
    Tok_T
    \right),
\end{equation}
where
\(
    \mathbf{Z}
    \in
    \mathbb{R}^{(N \cdot H) \times C},
\)
with \(H = 1\). Therefore, for each input sample in this study, the final output of the feature extraction module has the shape
\(
    \mathbf{Z}
    \in
    \mathbb{R}^{N\times C}.
\)

\subsection{Context Encoder}
After the Feature Extraction module, the fused token sequence
\(\mathbf{Z}\in\mathbb{R}^{N\times C}\) is passed to the Context
Encoder to model temporal dependencies among the chunk-level
representations. The Context Encoder consists of a GRU followed by
Causal Cosine Attention, which progressively transforms the compact
feature tokens into a historical context representation for SOC
estimation.

\paragraph{GRU}
The fused token sequence \(\mathbf{Z}\) is passed into a GRU encoder to capture temporal dependencies among chunk-level representations. 
% Given the input token \(\mathbf{z}_n\) at step \(n\) and the previous hidden state \(\mathbf{h}_{n-1}\), the GRU updates its hidden state using reset and update gates:
% \begin{equation}
%     \mathbf{r}_n
%     =
%     \sigma
%     \left(
%     \mathbf{W}_{r}\mathbf{z}_n
%     +
%     \mathbf{U}_{r}\mathbf{h}_{n-1}
%     +
%     \mathbf{b}_{r}
%     \right),
% \end{equation}
% \begin{equation}
%     \mathbf{u}_n
%     =
%     \sigma
%     \left(
%     \mathbf{W}_{u}\mathbf{z}_n
%     +
%     \mathbf{U}_{u}\mathbf{h}_{n-1}
%     +
%     \mathbf{b}_{u}
%     \right),
% \end{equation}
% \begin{equation}
%     \widetilde{\mathbf{h}}_n
%     =
%     \tanh
%     \left(
%     \mathbf{W}_{h}\mathbf{z}_n
%     +
%     \mathbf{U}_{h}
%     \left(
%     \mathbf{r}_n \odot \mathbf{h}_{n-1}
%     \right)
%     +
%     \mathbf{b}_{h}
%     \right),
% \end{equation}
% \begin{equation}
%     \mathbf{h}_n
%     =
%     \left(1-\mathbf{u}_n\right)
%     \odot
%     \mathbf{h}_{n-1}
%     +
%     \mathbf{u}_n
%     \odot
%     \widetilde{\mathbf{h}}_n,
% \end{equation}
% where \(\mathbf{r}_n\) is the reset gate, \(\mathbf{u}_n\) is the update gate, \(\widetilde{\mathbf{h}}_n\) is the candidate hidden state, \(\sigma(\cdot)\) is the sigmoid activation function, and \(\odot\) denotes element-wise multiplication. 
The initial hidden state is set to a zero vector:
\(
    \mathbf{h}_0 = \mathbf{0}.
\)
The output hidden sequence of the GRU encoder is denoted as
\begin{equation}
    \mathbf{H}
    =
    \text{GRU}
    \left(
    \mathbf{Z}
    \right)
    =
    \left[
    \mathbf{h}_1,
    \mathbf{h}_2,
    \ldots,
    \mathbf{h}_{N \cdot H}
    \right],
\end{equation}
where \(\mathbf{h}_n \in \mathbb{R}^{d}\), and \(d\) is the hidden dimension of the GRU encoder. The output hidden sequence has the shape
\(
    \mathbf{H}
    \in
    \mathbb{R}^{N \times d}.
\)

\paragraph{Causal Cosine Attention}
To further refine the temporal context, Causal Cosine Attention is applied to the GRU hidden sequence
\(
    \mathbf{H} \in \mathbb{R}^{N \times d}
\).
This module is to emphasize the most relevant historical hidden states based on cosine similarity while preserving causality. Given the GRU hidden states
\(
    \mathbf{H} = [\mathbf{h}_1,\mathbf{h}_2,\ldots,\mathbf{h}_{N}]
\),
each hidden state is first normalized, and the cosine similarity between time steps \(i\) and \(j\) is computed as
\begin{equation}
    \widetilde{\mathbf{h}}_i
    =
    \frac{\mathbf{h}_i}{\left\|\mathbf{h}_i\right\|_2},
    \qquad
    s_{ij}
    =
    \widetilde{\mathbf{h}}_i^{\top}
    \widetilde{\mathbf{h}}_j.
\end{equation}
To ensure causal modeling, future time steps are masked before the softmax operation:
\begin{equation}
    \bar{s}_{ij}
    =
    \begin{cases}
        s_{ij}, & j \leq i, \\
        -\infty, & j > i.
    \end{cases}
\end{equation}
The causal attention weight and refined context representation are then computed as
\begin{equation}
    a_{ij}
    =
    \frac{
    \exp
    \left(
    \bar{s}_{ij}/\tau
    \right)
    }
    {
    \sum_{k \leq i}
    \exp
    \left(
    \bar{s}_{ik}/\tau
    \right)
    },
    \qquad
    \mathbf{g}_i
    =
    \sum_{j \leq i}
    a_{ij}
    \mathbf{h}_j,
\end{equation}
where \(\tau\) is the temperature parameter controlling the sharpness of the attention distribution. 

The output of the Causal Cosine Attention module is
\(
    \mathbf{G}
    =
    [
    \mathbf{g}_1,
    \mathbf{g}_2,
    \ldots,
    \mathbf{g}_{N}
    ]
    \in
    \mathbb{R}^{N \times d}.
\)
The final historical context vector is selected from the last time step:
\(
    \mathbf{g}_{t}
    =
    \mathbf{g}_{N}
    \in
    \mathbb{R}^{d}.
\)
This summarizes all available causal information from the short input window up to the current estimation time, and is used as the contextual state for the next module.

\subsection{Latest-Measurement Decoder}
After the Context Encoder, the final contextual state \(\mathbf{g}_t\) summarizes the relevant historical information from
the short observation window. The Latest-Measurement Decoder combines
this contextual state with the latest measurement
\(\mathbf{x}_t=[I_t,V_t,T_t]\) to update the representation according
to the current battery operating condition and produce the final SOC
estimate.

\paragraph{GRUCell}
The latest-measurement decoder is inspired by a Kalman-like state update mechanism~\cite{welch1995introduction}, where the contextual state is refined using the latest measurement:
\(
    \mathbf{x}_t = [I_t, V_t, T_t].
\)
The update is performed using a GRUCell:
\begin{equation}
    \mathbf{h}^{dec}_t
    =
    \mathbf{g}_t
    +
    \Delta
    \left(
    \mathbf{x}_t, \mathbf{g}_t
    \right)
     =
    \mathrm{GRUCell}
    \left(
    \mathbf{x}_t,
    \mathbf{g}_t
    \right),
\end{equation}
where \(\mathbf{g}_t \in \mathbb{R}^{d}\) is the contextual state from the encoder, \(\mathbf{h}^{dec}_t \in \mathbb{R}^{d}\) is the latest-aware hidden state, and \(\Delta(\cdot)\) denotes a learnable correction function. In this formulation, the GRUCell acts as a learnable filtering function that updates the historical context using the current battery operating condition.

\paragraph{MLP}
The latest-aware hidden state is passed into an MLP head:
\begin{equation}
    \widehat{SOC}_t
    =
    \sigma
    \left(
    \mathbf{W}_{o}
    \mathrm{Dropout}
    \left(
    \mathrm{LayerNorm}
    \left(
    \mathbf{h}^{dec}_t
    \right)
    \right)
    +
    b_o
    \right),
\end{equation}
where \(\sigma(\cdot)\) is the sigmoid activation function. The sigmoid function constrains the SOC prediction to the range \([0,1]\). A summary of the proposed C2L-Net architecture is provided in Table~\ref{tab:c2lnet_architecture}.

\begin{table*}[h]
\centering
\caption{Architecture details of the proposed C2L-Net model.}
\label{tab:c2lnet_architecture}
\resizebox{1\textwidth}{!}{%
\begin{tabular}{llll}
\toprule
\textbf{Module} & \textbf{Layers} & \textbf{Input Shape} & \textbf{Output Shape} \\
\midrule
Chunking 
& Reshape into chunks 
& $(B, 200, 3)$ 
& $(B, 5, 40, 3)$ \\

\midrule
\multirow{3}{*}{Feature Extraction}
& Theta Attention Pool 
& $(B, 5, 40)$ 
& $(B, 5, 21)$ \\
& Seasonality Basis (Fourier) 
& $(B, 5, 21)$ 
& $(B, 5, 1)$ \\
& Token stacking (I, V, T) 
& $(B, 5, 1) \times 3$ 
& $(B, 5, 3)$ \\

\midrule
\multirow{3}{*}{Context Encoder}
& GRU 
& $(B, 5, 3)$ 
& $(B, 5, 128)$ \\
& Causal Cosine Attention 
& $(B, 5, 128)$ 
& $(B, 5, 128)$ \\
& State vector (last step) 
& $(B, 5, 128)$ 
& $(B, 128)$ \\

\midrule
\multirow{2}{*}{Latest-Measurement Decoder}
& GRUCell (latest input + State) 
& $(B, 3), (B, 128)$ 
& $(B, 128)$ \\
& MLP Head (LN--Dropout--Linear--Sigmoid) 
& $(B, 128)$ 
& $(B, 1)$ \\

\bottomrule
\end{tabular}
}
\end{table*}

The proposed C2L-Net has several advantages for realistic online SOC estimation. First, it uses only a short measurement window, avoiding dependence on long historical profiles and reducing zero-padding bias at the initial stage. Second, the chunking and feature extraction modules capture local temporal patterns from current, voltage, and temperature separately, while shortening the temporal dimension to reduce the computational cost of the subsequent state-space modules and enable faster inference. Third, the causal temporal cosine attention module refines the contextual state while preventing future information leakage. Finally, the latest-measurement decoder updates the contextual state using the latest measurement, enabling more accurate SOC estimation based on both recent historical context and the current battery operating condition.

\section{Dataset}\label{Dataset}
\paragraph{Data Description}
We use the public lithium-ion battery drive-cycle dataset introduced by \cite{yao2025multi} to evaluate the proposed model for SOC estimation. This dataset was designed for online SOC estimation and contains dynamic battery cycling profiles under different driving patterns and ambient temperature conditions. The dataset was collected using LG INR 21700 M50LT lithium-ion cells with nickel manganese cobalt (NMC) chemistry. The nominal capacity and nominal energy of the cell are 4.93 Ah and 18.2 Wh, respectively. The nominal voltage is 3.69 V, with a maximum charge voltage of 4.20 V and a minimum discharge voltage of 2.50 V. The dataset includes twelve distinct drive cycles: Braunschweig City Driving Cycle (BCDC), California Unified Cycle (LA92), Heavy Heavy-Duty Diesel Truck Composite Cycle (HHDDT), City Suburban Heavy Vehicle Cycle (CSHVC), Federal Test Procedure-72 (FTP-72), Federal Test Procedure-75 (FTP-75), Highway Fuel Economy Test (HWFET), Inspection and Maintenance (IM), US06 Supplemental Federal Test Procedure (US06), Parcel Delivery Truck Cycle Baltimore (PDTCB), Port Drayage Metro Highway Cycle California (PDMHC), and Orange County Transit Bus Cycle (OCTBC). These drive cycles cover different usage scenarios and cycling patterns, ranging from light-duty to heavy-duty driving conditions. The battery tests were conducted under five fixed ambient temperatures:
\begin{equation}
    5^{\circ}\mathrm{C}, \quad
    15^{\circ}\mathrm{C}, \quad
    25^{\circ}\mathrm{C}, \quad
    35^{\circ}\mathrm{C}, \quad
    45^{\circ}\mathrm{C}.
\end{equation}
For each temperature condition, the battery was first charged using a constant-current/constant-voltage (CC-CV) charging protocol until fully charged. After a relaxation period, the battery was repeatedly cycled using the corresponding drive-cycle power profile until the terminal voltage reached the lower cut-off voltage. The measured voltage, current, and temperature signals were then used to construct input samples, while the corresponding SOC value was used as the prediction target.

Following the original dataset setting, the drive cycles are divided into training, validation, and testing groups. The training set contains BCDC, LA92, CSHVC, HWFET, IM, US06, PDTCB, and OCTBC. The validation set contains HHDDT and FTP-72, while the test set contains FTP-75 and PDMHC. This split ensures that the model is evaluated on drive cycles that are not used during training, allowing a more reliable assessment of generalization performance.

\begin{figure}[H]
    \centering
    \includegraphics[width=\textwidth]{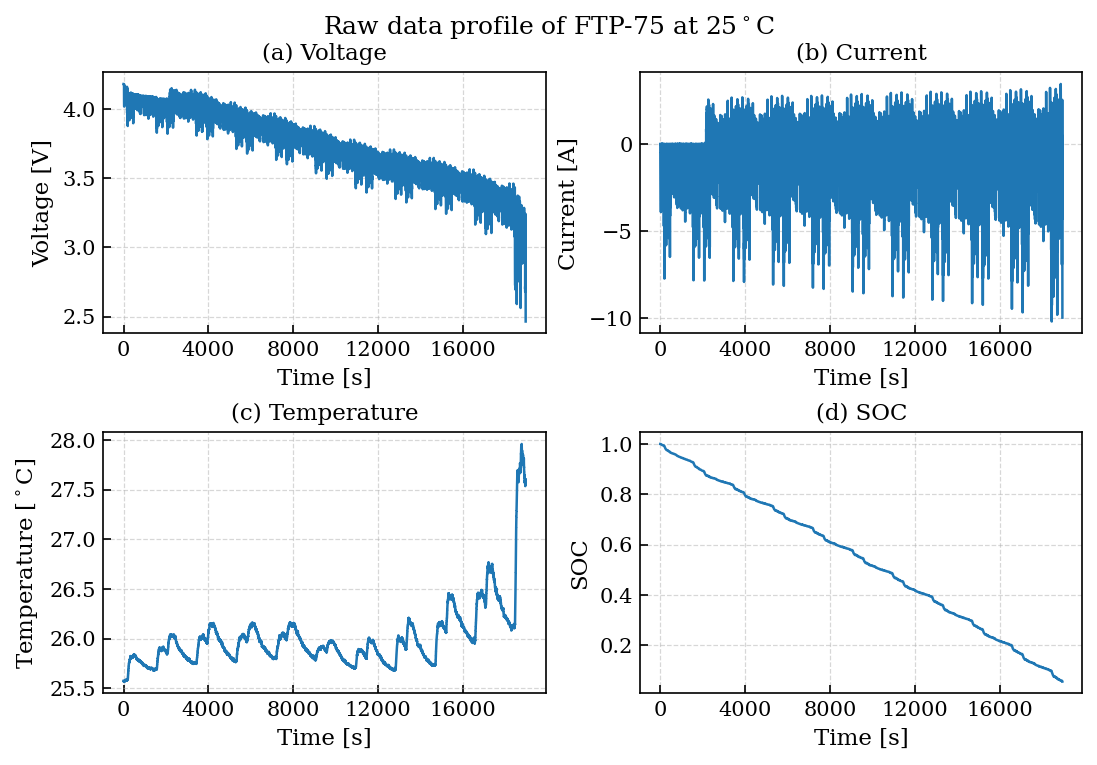}
    \caption{Raw data profile of the FTP-75 drive cycle at \(25^\circ\mathrm{C}\).}
    \label{fig:ftp75_25deg_profile}
\end{figure}

\paragraph{Data Analysis}
Fig.~\ref{fig:ftp75_25deg_profile} shows the raw data profile of the FTP-75 drive cycle at \(25^\circ\mathrm{C}\), including voltage, current, temperature, and SOC. The voltage gradually decreases from approximately \(4.1~\mathrm{V}\) to around \(2.5~\mathrm{V}\), reflecting the battery discharge process under the dynamic drive-cycle load. A sharp voltage drop is observed near the end of the profile, which indicates that the battery approaches the low-SOC region where voltage changes more rapidly.
The current profile shows frequent fluctuations caused by the FTP-75 driving pattern, where negative values represent discharge events and positive values indicate regenerative or charging-like events. These rapid current changes lead to transient voltage responses, while the temperature gradually increases from around \(25^\circ\mathrm{C}\) due to internal heat generation during operation. The SOC decreases from nearly \(1.0\) to close to \(0\), representing the overall discharge process. This profile illustrates a realistic driving condition, where frequent load variations and nonlinear battery responses require accurate and robust SOC estimation.

\section{Experiments and Discussion}\label{Experiments_Discussion}

\subsection{Experimental Setup}
All experiments were conducted on a desktop computer equipped with an NVIDIA RTX 3060 GPU, an AMD Ryzen 7 processor, and 32 GB of RAM. The proposed C2L-Net was implemented using PyTorch and trained for 100 epochs. The AdamW optimizer~\cite{meng2023machine} was used to update the model parameters with a learning rate of \(5 \times 10^{-4}\). The mean squared error (MSE) loss function was adopted as the training objective, which is defined as
\begin{equation}
    \mathcal{L}_{\mathrm{MSE}}
    =
    \frac{1}{M}
    \sum_{i=1}^{M}
    \left(
    SOC_i - \widehat{SOC}_i
    \right)^2,
\end{equation}
where \(M\) denotes the batch size, which was set to 128 for all experiments, \(SOC_i\) is the ground-truth SOC, and \(\widehat{SOC}_i\) is the predicted SOC. Following~\cite{yao2025multi}, Min--Max normalization was applied to the input features. The scaler was fitted using only the training data and then used to transform the validation and testing data. These settings were kept consistent across all experiments to ensure a fair evaluation of the proposed model.

The model performance was evaluated using three metrics: mean absolute error (MAE), root mean squared error (RMSE), and maximum absolute error (MAX), which are defined as follows:
\begin{equation}
    \mathrm{MAE}
    =
    \frac{1}{n}
    \sum_{i=1}^{n}
    \left|
    SOC_i - \widehat{SOC}_i
    \right|,
\end{equation}
\begin{equation}
    \mathrm{RMSE}
    =
    \sqrt{
    \frac{1}{n}
    \sum_{i=1}^{n}
    \left(
    SOC_i - \widehat{SOC}_i
    \right)^2
    },
\end{equation}
\begin{equation}
    \mathrm{MAX}
    =
    \max_{1 \leq i \leq n}
    \left|
    SOC_i - \widehat{SOC}_i
    \right|,
\end{equation}
where \(n\) is the number of test samples. All experiments were repeated three times with different random seeds, and the reported results correspond to the average performance.

\subsection{Evaluation}
We first conduct an ablation study using the $5^\circ$C data to identify the best model configuration. Then, we compare the selected C2L-Net configuration with existing models under fixed-temperature conditions for overall validation. Finally, we perform a computational efficiency analysis to evaluate its suitability for real-world deployment.

\paragraph{Ablation Study of Model Components}
Table~\ref{tab:ablation_each_block} presents the ablation results obtained by replacing the base components in each module with alternative configurations to evaluate their effectiveness. The alternative configurations considered in this study include the Temporal Convolutional Network (TCN)~\cite{lea2016temporal}, multi-head self-attention~\cite{vaswani2017attention}, Cottention~\cite{mongaras2025cottention}, and Long Short-Term Memory (LSTM)~\cite{graves2012long}. For feature extraction, the proposed Theta Attention Pool + Seasonality Basic outperforms the TCN-based alternative, reducing the average MAE from $1.5856\%$ to $1.0386\%$ and the average RMSE from $2.0634\%$ to $1.3299\%$. This indicates that the proposed chunk-wise feature extraction module is more effective for capturing local battery dynamics. For the encoder--decoder design, all configurations include an MLP head to produce the final SOC estimation. Encoder-only architectures generally yield higher errors, particularly multi-head self-attention and cosine similarity-based attention, which achieve average MAE values of $3.3816\%$ and $5.7616\%$, respectively. In contrast, encoder--decoder variants achieve lower errors, highlighting the benefit of updating the historical context with the latest measurement. Among all configurations, the proposed GRU--Cosine Attention--GRUCell base model achieves an average MAE of $1.0386\%$, RMSE of $1.3299\%$, and MAX error of $7.3633\%$. It also maintains stable performance across both FTP-75 and PDMHC cycles, validating the effectiveness of combining chunk-based feature extraction, causal context modeling, and a Kalman-like latest-measurement update for robust SOC estimation.

\begin{table*}[h]
\centering
\caption{Ablation study of SOC prediction performance under different module configurations at $5^\circ$C.}
\label{tab:ablation_each_block}
\resizebox{\textwidth}{!}{%
\begin{tabular}{lllccccccccc}
\toprule
\textbf{Module} & \textbf{Type} & \textbf{Configuration} &
\multicolumn{3}{c}{\textbf{FTP-75}} &
\multicolumn{3}{c}{\textbf{PDMHC}} &
\multicolumn{3}{c}{\textbf{Average}} \\
\cmidrule(lr){4-6}\cmidrule(lr){7-9}\cmidrule(lr){10-12}
& & &
\textbf{MAE [\%]} & \textbf{RMSE [\%]} & \textbf{MAX [\%]} &
\textbf{MAE [\%]} & \textbf{RMSE [\%]} & \textbf{MAX [\%]} &
\textbf{MAE [\%]} & \textbf{RMSE [\%]} & \textbf{MAX [\%]} \\
\midrule

\multirow{2}{*}{\textbf{Feature Extraction}}
& Proposed 
& \textbf{Theta Attention Pool + Seasonality Basic \textit{(base model)}}
& \textbf{0.9744} & \textbf{1.2238} & \textbf{7.1213}
& \textbf{1.1027} & \textbf{1.4359} & \textbf{7.6053}
& \textbf{1.0386} & \textbf{1.3299} & \textbf{7.3633} \\

& Alternative 
& TCN
& 1.6598 & 2.1164 & 10.7045
& 1.5114 & 2.0103 & 7.0261
& 1.5856 & 2.0634 & 8.8653 \\

\midrule

\multirow{9}{*}{\textbf{Encoder--Decoder}}
& Encoder only
& Multi-head self-attention
& 4.0565 & 6.4077 & 22.1911
& 2.7066 & 3.4717 & 11.6698
& 3.3816 & 4.9397 & 16.9305 \\

& Encoder only
& Cosine similarity-based attention
& 7.4948 & 11.4479 & 29.8436
& 4.0283 & 5.8291 & 17.0869
& 5.7616 & 8.6385 & 23.4653 \\

& Encoder only
& Cottention
& 1.4673 & 1.8708 & 8.6399
& 1.7299 & 2.1445 & 7.1025
& 1.5986 & 2.0077 & 7.8712 \\

& Encoder only
& GRU
& 1.3113 & 1.5608 & 7.2552
& 1.0775 & 1.3622 & 7.1398
& 1.1944 & 1.4615 & 7.1975 \\

& Encoder only
& LSTM
& 1.8187 & 2.3056 & 9.4995
& 1.1265 & 1.4356 & 6.4584
& 1.4726 & 1.8706 & 7.9790 \\

& Encoder--Decoder
& GRU--GRUCell
& 0.6714 & 0.9064 & 6.0898
& 1.4870 & 1.9539 & 10.9024
& 1.0792 & 1.4302 & 8.4961 \\

& Encoder--Decoder
& LSTM--LSTMCell
& 0.7979 & 0.9884 & 7.1566
& 1.4401 & 1.9114 & 7.0471
& 1.1190 & 1.4499 & 7.1019 \\

& Encoder--Decoder
& \textbf{GRU--Cosine Attention--GRUCell \textit{(base model)}}
& \textbf{0.9744} & \textbf{1.2238} & \textbf{7.1213}
& \textbf{1.1027} & \textbf{1.4359} & \textbf{7.6053}
& \textbf{1.0386} & \textbf{1.3299} & \textbf{7.3633} \\

& Encoder--Decoder
& LSTM--Cosine Attention--LSTMCell
& 0.7096 & 0.9401 & 6.1947
& 1.6483 & 2.2159 & 8.3393
& 1.1790 & 1.5780 & 7.2670 \\

\bottomrule
\end{tabular}
}
\end{table*}

\begin{figure}[h]
    \centering
    \includegraphics[width=\textwidth]{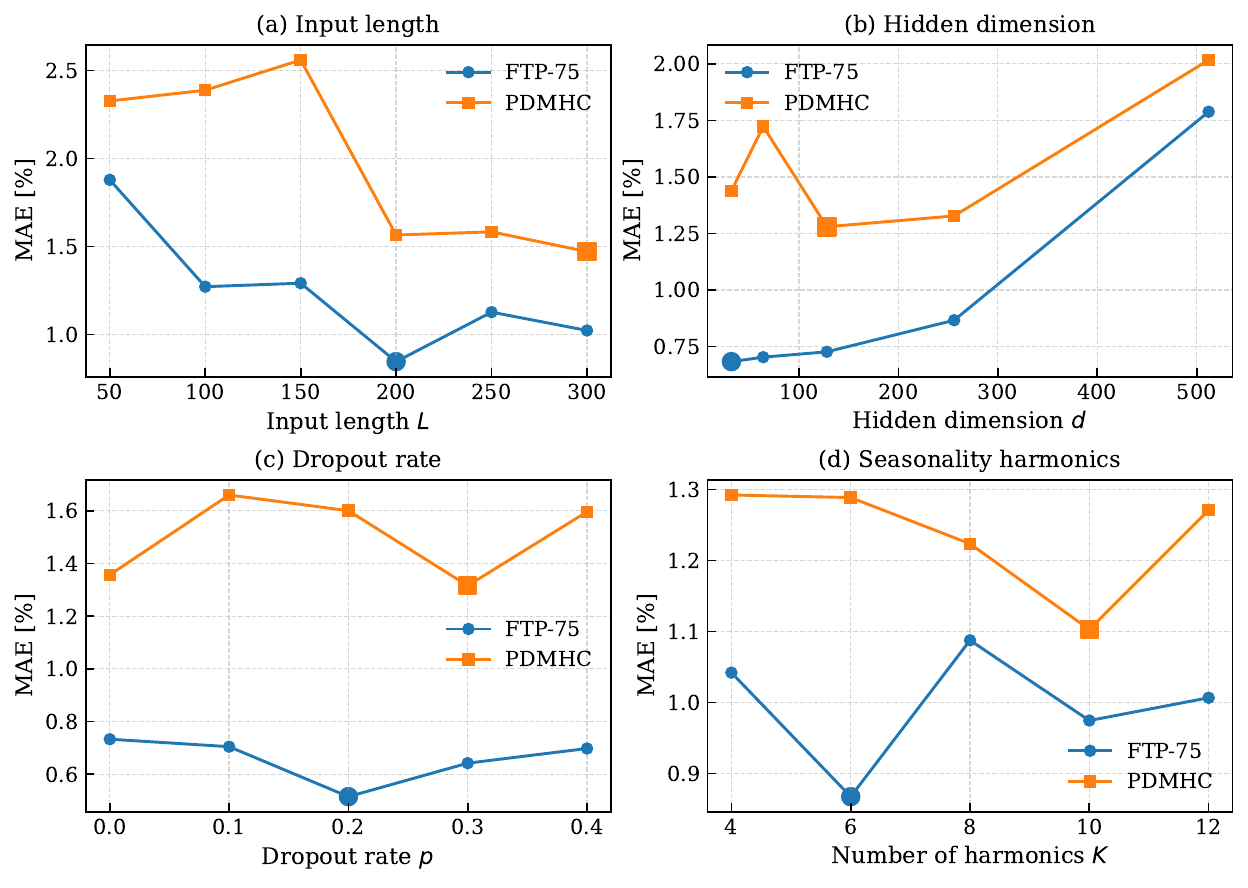}
    \caption{MAE comparison of the proposed C2L-Net under different hyperparameter settings, including input length, hidden dimension, dropout rate, and seasonality harmonics.}
    \label{fig:ablation_mae_summary}
\end{figure}
Fig.~\ref{fig:ablation_mae_summary} summarizes the MAE trends of C2L-Net under different hyperparameter settings on the FTP-75 and PDMHC test profiles. For the input length, very short windows such as \(L=50\) and \(L=100\) lead to larger errors because they provide limited historical context, while \(L=200\) achieves the best overall trade-off between accuracy and short-window practicality. For the hidden dimension, increasing \(d\) does not consistently improve performance; in particular, \(d=512\) produces higher errors, suggesting possible over-parameterization and weaker generalization. The dropout analysis shows that moderate regularization is beneficial, with \(p=0.2\) giving the lowest MAE on FTP-75 and competitive performance on PDMHC. For the Seasonality Basis, a moderate number of harmonics provides better performance than too few or too many harmonics, and \(K=10\) achieves the best MAE on PDMHC while remaining competitive on FTP-75. Based on these observations, the final configuration of C2L-Net is selected as \(L=200\), \(d=128\), \(p=0.2\), and \(K=10\), which provides a balanced choice between estimation accuracy, computational efficiency, and online applicability.

\paragraph{Performance Comparison with Baseline Models}

Table~\ref{tab:avg_temp_best} compares the SOC prediction performance of the proposed C2L-Net with several baseline models across five fixed ambient temperatures. The baseline models include fundamental architectures such as TCN, LSTM, GRU, and Transformer encoder, as well as recent methods such as TTSNet~\cite{bao2024ttsnet} and TCN-Short~\cite{yao2025multi}. For TTSNet, we re-ran the model under our experimental settings while preserving its original architecture to ensure a fair comparison. For TCN-Short~\cite{yao2025multi}, its short receptive field of approximately 0.8~s reduces its dependence on padding, making it a relevant baseline for comparison. At \(45^\circ\mathrm{C}\), the proposed model achieves the best average performance, with an average MAE of \(0.4118\%\), RMSE of \(0.5320\%\), and MAX error of \(2.4923\%\), outperforming LSTM, which obtains an average MAE of \(0.4191\%\), RMSE of \(0.5585\%\), and MAX error of \(3.8197\%\). Although LSTM performs best on FTP-75 at this temperature, C2L-Net achieves the best results on PDMHC with an MAE of \(0.5512\%\), RMSE of \(0.6860\%\), and MAX error of \(2.9189\%\), indicating stronger generalization across unseen drive cycles. At \(35^\circ\mathrm{C}\), LSTM obtains the best average MAE and RMSE, with \(0.6666\%\) and \(0.8892\%\), respectively, while the proposed model achieves an average MAE of \(0.7529\%\) and RMSE of \(1.0363\%\). This suggests that C2L-Net is slightly less effective than LSTM under this temperature condition, but still remains competitive with other baselines such as GRU, Transformer encoder, TCN-Short, and TTSNet. At \(25^\circ\mathrm{C}\), the proposed C2L-Net achieves the best average MAE and RMSE, with \(0.6708\%\) and \(0.9576\%\), respectively. It also achieves the best FTP-75 performance, with an MAE of \(0.4055\%\), RMSE of \(0.5896\%\), and MAX error of \(3.3789\%\), and the best PDMHC MAE and RMSE, with \(0.9360\%\) and \(1.3255\%\), respectively. These results show that the proposed short-window context representation can effectively capture the relationship between voltage, current, temperature, and SOC under moderate ambient temperature. At \(15^\circ\mathrm{C}\), C2L-Net achieves the best average MAE and RMSE, with \(0.7529\%\) and \(1.0363\%\), respectively, while the Transformer encoder and LSTM show slightly better performance on some individual metrics. This indicates that the proposed model provides a more balanced overall performance across both FTP-75 and PDMHC rather than being optimized for only one test profile. At the low-temperature condition of \(5^\circ\mathrm{C}\), the proposed model again achieves the best average MAE and RMSE, with \(1.0386\%\) and \(1.3299\%\), respectively, outperforming Transformer encoder, TTSNet, GRU, LSTM, TCN, and TCN-Short in terms of average accuracy. In particular, C2L-Net obtains the best PDMHC performance at \(5^\circ\mathrm{C}\), with an MAE of \(1.1027\%\) and RMSE of \(1.4359\%\), showing its robustness under more challenging low-temperature dynamics. Overall, C2L-Net achieves the best average MAE and RMSE at four out of five temperature conditions \((45^\circ\mathrm{C}, 25^\circ\mathrm{C}, 15^\circ\mathrm{C},\) and \(5^\circ\mathrm{C})\), demonstrating that the proposed architecture provides strong and stable SOC estimation performance across different thermal environments. The results also show that TCN-based models, especially TCN-Short, can suffer from large MAX errors under some conditions, while the proposed C2L-Net generally maintains lower average errors by combining short-window feature extraction, recurrent context encoding, causal cosine attention, and latest-measurement decoding. These findings validate the effectiveness of the proposed design for realistic online SOC estimation under fixed ambient temperature conditions. The strong performance of the proposed model can be attributed to the combination of compact temporal representation and explicit latest-measurement updating. The chunk-based encoding captures essential short-term dynamics, while the decoder mechanism ensures rapid adaptation to instantaneous changes in operating conditions, which are critical for accurate SOC estimation under dynamic loads.

\begin{table*}[h]
\centering
\caption{SOC prediction performance with averaged results across FTP-75 and PDMHC. Best results per temperature and metric are in bold.}
\label{tab:avg_temp_best}
\resizebox{1\textwidth}{!}{%
\begin{tabular}{c l c c c c c c c c c}
\toprule
\multirow{2}{*}{\textbf{Temp. [$^\circ$C]}} & \multirow{2}{*}{\textbf{Model}} &
\multicolumn{3}{c}{\textbf{FTP-75}} &
\multicolumn{3}{c}{\textbf{PDMHC}} &
\multicolumn{3}{c}{\textbf{Average}} \\
\cmidrule(lr){3-5}\cmidrule(lr){6-8}\cmidrule(lr){9-11}
& & MAE & RMSE & MAX & MAE & RMSE & MAX & MAE & RMSE & MAX \\
\midrule

\multirow{7}{*}{45}
& TCN  & 0.3888 & 0.5232 & 3.5769 & 1.652 & 2.5447 & 13.9808 & 1.0204 & 1.5339 & 8.7789 \\
& LSTM & \textbf{0.2327} & \textbf{0.3269} & \textbf{1.9538} & 0.6054 & 0.79 & 5.6855 & 0.4191 & 0.5585 & 3.8197 \\
& GRU  & 0.7328 & 0.9512 & 5.6871 & 1.3568 & 1.7348 & 6.461 & 1.0448 & 1.3430 & 6.0741 \\
& Transformer encoder & 0.3523 & 0.442 & 2.5327 & 0.7106 & 0.8785 & 4.1883 & 0.5315 & 0.6603 & 3.3605 \\
& TCN-Short & 0.671 & 0.925 & 5.490 & 1.046 & 1.317 & 5.857 & 0.8585 & 1.1210 & 5.6735 \\
& TTSNet & 0.3448 & 0.4696 & 2.7997 & 0.638 & 0.8246 & 4.7667 & 0.4914 & 0.6471 & 3.7832 \\
& Ours & 0.2723 & 0.378 & 2.0656 & \textbf{0.5512} & \textbf{0.686} & \textbf{2.9189} & \textbf{0.4118} & \textbf{0.5320} & \textbf{2.4923} \\
\midrule

\multirow{7}{*}{35}
& TCN  & 0.595 & 0.7542 & 3.8662 & 2.4864 & 3.2747 & 11.2047 & 1.5407 & 2.0145 & 7.5355 \\
& LSTM & 0.5372 & 0.7313 & 3.5796 & \textbf{0.7959} & \textbf{1.0471} & \textbf{4.9303} & \textbf{0.6666} & \textbf{0.8892} & \textbf{4.2550} \\
& GRU  & \textbf{0.4772} & \textbf{0.6592} & 4.1564 & 0.8712 & 1.1668 & 5.1096 & 0.6742 & 0.9130 & 4.6330 \\
& Transformer encoder & 0.5304 & 0.6824 & \textbf{3.4088} & 0.8776 & 1.1012 & 4.497 & 0.7040 & 0.8918 & 3.9529 \\
& TCN-Short & 0.831 & 1.190 & 23.264 & 1.469 & 1.862 & 23.110 & 1.1500 & 1.5260 & 23.1870 \\
& TTSNet & 0.6128 & 0.7825 & 3.529 & 0.8332 & 1.2251 & 5.5996 & 0.7230 & 1.0038 & 4.5643 \\
& Ours & 0.5496 & 0.7462 & 5.2353 & 0.9561 & 1.3263 & 7.3945 & 0.7529 & 1.0363 & 6.3149 \\
\midrule

\multirow{7}{*}{25}
& TCN  & 1.3145 & 1.8096 & 10.1419 & 2.1664 & 3.0731 & 13.9676 & 1.7405 & 2.4414 & 12.0548 \\
& LSTM & 0.6019 & 0.8121 & 5.3082 & 1.1804 & 1.5203 & 6.5734 & 0.8912 & 1.1662 & 5.9408 \\
& GRU  & 0.5833 & 0.7432 & 5.4428 & 1.247 & 1.8049 & 9.2982 & 0.9152 & 1.2741 & 7.3705 \\
& Transformer encoder & 0.5521 & 0.7186 & 4.6819 & 1.2377 & 1.6142 & \textbf{5.3759} & 0.8949 & 1.1664 & \textbf{5.0289} \\
& TCN-Short & 0.974 & 1.380 & 54.062 & 1.785 & 2.319 & 54.087 & 1.3795 & 1.8495 & 54.0745 \\
& TTSNet & 0.4806 & 0.7107 & 4.4645 & 1.1329 & 1.5744 & 8.2804 & 0.8068 & 1.1426 & 6.3725 \\
& Ours & \textbf{0.4055} & \textbf{0.5896} & \textbf{3.3789} & \textbf{0.936} & \textbf{1.3255} & 9.3854 & \textbf{0.6708} & \textbf{0.9576} & 6.3822 \\
\midrule

\multirow{7}{*}{15}
& TCN  & 2.0535 & 2.6526 & 13.073 & 2.1287 & 3.0091 & 12.4231 & 2.0911 & 2.8309 & 12.7481 \\
& LSTM & 0.6524 & 0.89 & \textbf{4.1034} & 1.2232 & 1.6953 & 7.7804 & 0.9378 & 1.2927 & \textbf{5.9419} \\
& GRU  & \textbf{0.4698} & \textbf{0.6461} & 4.1299 & 1.2568 & 1.8896 & 8.9921 & 0.8633 & 1.2679 & 6.5610 \\
& Transformer encoder & 0.6088 & 0.7989 & 4.9604 & \textbf{1.092} & \textbf{1.4362} & 7.0273 & 0.8504 & 1.1176 & 5.9939 \\
& TCN-Short & 1.069 & 1.592 & 12.886 & 1.779 & 2.342 & 7.709 & 1.4240 & 1.9670 & 10.2975 \\
& TTSNet & 0.5663 & 0.8402 & 5.7459 & 1.2541 & 1.6004 & \textbf{6.5053} & 0.9102 & 1.2203 & 6.1256 \\
& Ours & 0.5496 & 0.7462 & 5.2353 & 0.9561 & 1.3263 & 7.3945 & \textbf{0.7529} & \textbf{1.0363} & 6.3149 \\
\midrule

\multirow{7}{*}{5}
& TCN  & 5.1506 & 6.953 & 19.2782 & 3.816 & 5.6463 & 19.0563 & 4.4833 & 6.2997 & 19.1673 \\
& LSTM & 1.3606 & 1.722 & 7.5379 & 1.6597 & 2.1164 & 7.465 & 1.5102 & 1.9192 & 7.5015 \\
& GRU  & 1.0119 & 1.3841 & 6.4654 & 1.9717 & 2.3862 & 7.6474 & 1.4918 & 1.8852 & 7.0564 \\
& Transformer encoder & \textbf{0.8072} & \textbf{1.113} & \textbf{5.7629} & 1.4576 & 1.7836 & \textbf{7.0913} & 1.1324 & 1.4483 & \textbf{6.4271} \\
& TCN-Short & 1.163 & 1.732 & 13.925 & 2.243 & 2.862 & 8.664 & 1.7030 & 2.2970 & 11.2945 \\
& TTSNet & 1.1508 & 1.4456 & 8.2374 & 1.6697 & 2.0404 & 8.2049 & 1.4103 & 1.7430 & 8.2212 \\
& Ours & 0.9744 & 1.2238 & 7.1213 & \textbf{1.1027} & \textbf{1.4359} & 7.6053 & \textbf{1.0386} & \textbf{1.3299} & 7.3633 \\
\bottomrule
\end{tabular}
}
\end{table*}

\paragraph{SOC Estimation and Error Analysis}

\begin{figure}[h]
\centering
\begin{subfigure}{0.48\textwidth}
    \centering
    \includegraphics[width=\linewidth]{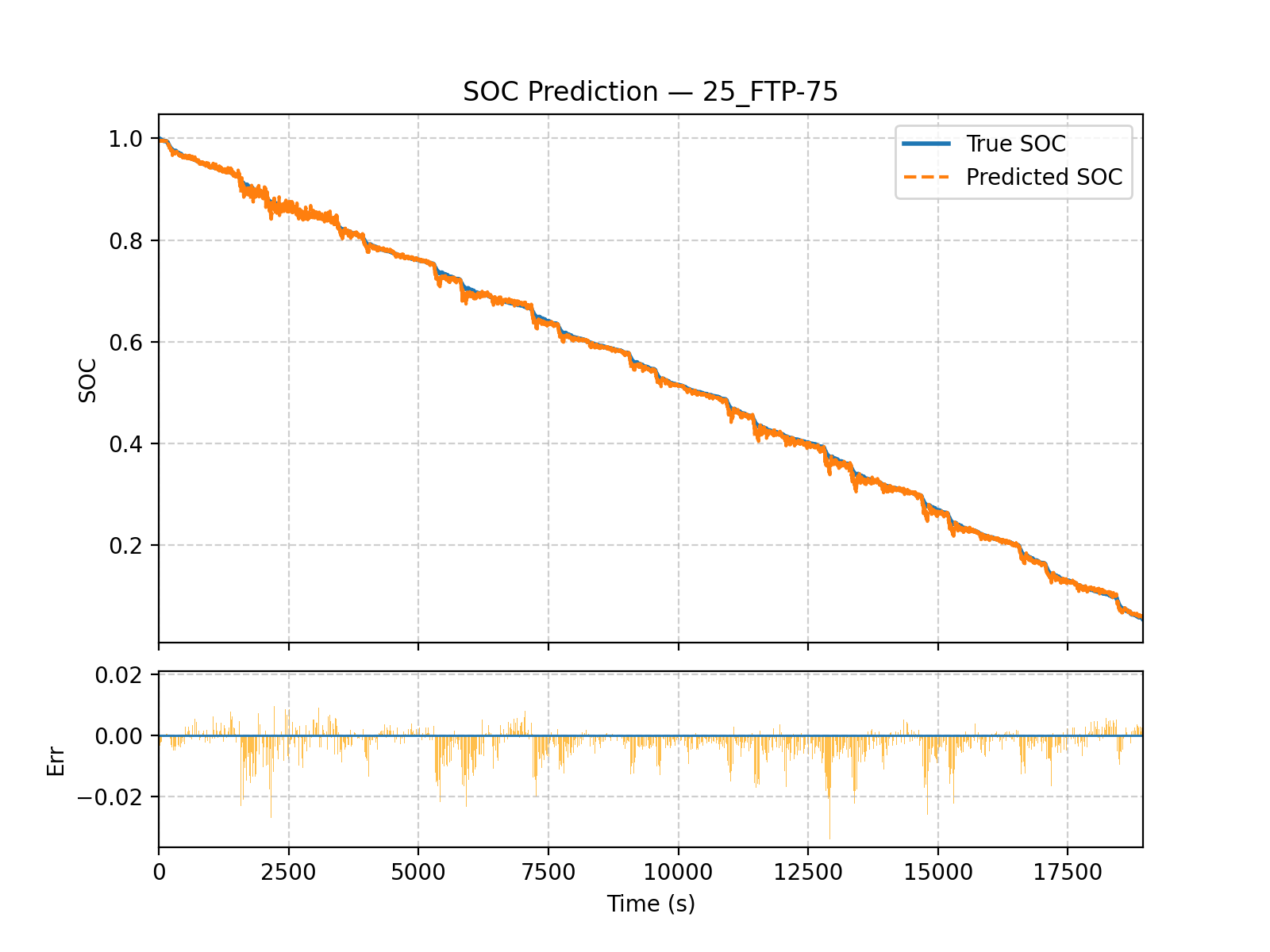}
    \caption{the FTP-75 drive cycle}
    \label{fig:ftp75_soc_error_25}
\end{subfigure}
\hfill
\begin{subfigure}{0.48\textwidth}
    \centering
    \includegraphics[width=\linewidth]{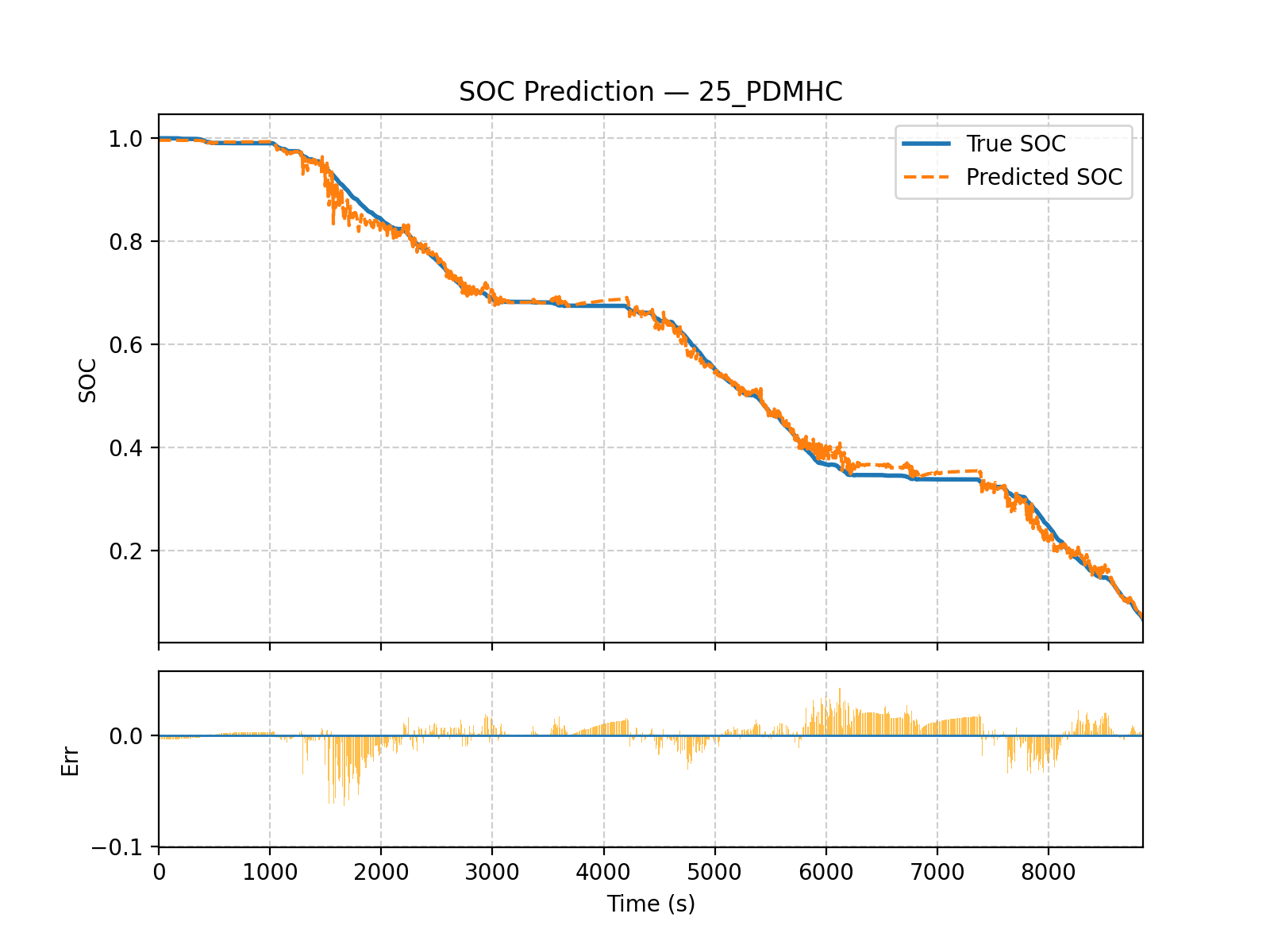}
    \caption{the PDMHC drive cycle}
    \label{fig:pdmhc_soc_error_25}
\end{subfigure}

\caption{SOC estimation error visualization on FTP-75 and PDMHC testing profiles at $25^\circ$C.}
\label{fig:soc_error_25}
\end{figure}

Fig.~\ref{fig:soc_error_25} demonstrate the effectiveness of the proposed C2L-Net model under dynamic drive-cycle conditions at $25^\circ$C. For both FTP-75 and PDMHC profiles, the predicted SOC closely follows the ground-truth trajectory throughout the entire discharge process, indicating that the model can accurately capture the nonlinear relationship between current, voltage, temperature, and SOC. The error plots show that most prediction errors are concentrated within a narrow range around zero, with only occasional spikes during rapid transient events, such as sudden load changes. These deviations are expected due to abrupt variations in operating conditions but remain small and quickly corrected, demonstrating the robustness of the model. C2L-Net maintains stable performance across both smooth and highly dynamic regions, confirming its ability to generalize across different driving patterns.

\paragraph{Computational Efficiency Analysis}
Table~\ref{tab:computational_efficiency_comparison} compares the
computational efficiency of the proposed C2L-Net with TCN-Short and
TTSNet. The experiments are conducted on a desktop computer equipped
with an NVIDIA RTX 3060 GPU, an AMD Ryzen 7 processor, and \(32~\mathrm{GB}\)
of RAM. All models are implemented in PyTorch, and inference is
performed using the CPU. The proposed C2L-Net achieves the lowest
computational cost, with only 161,347 parameters and a model size of
\(0.62~\mathrm{MB}\), both smaller than those of TCN-Short
(177,249 parameters, \(0.68~\mathrm{MB}\)) and TTSNet
(1,657,988 parameters, \(6.32~\mathrm{MB}\)). In terms of inference
latency, C2L-Net significantly outperforms the baseline models, achieving
a mean latency of \(0.30~\mathrm{ms}\), compared with
\(18.68~\mathrm{ms}\) for TCN-Short and \(12.22~\mathrm{ms}\) for
TTSNet. This corresponds to a throughput of 3368.4 inferences/s,
making C2L-Net more than \(60\times\) faster than TCN-Short and over
\(40\times\) faster than TTSNet. These results demonstrate that C2L-Net
provides a highly efficient architecture for real-time SOC estimation,
combining low computational complexity with fast inference. Therefore,
the proposed model is particularly suitable for deployment in
resource-constrained BMS and edge devices.
\begin{table*}[h]
\centering
\caption{Computational efficiency comparison between C2L-Net and baseline models.}
\label{tab:computational_efficiency_comparison}
\resizebox{\columnwidth}{!}{%
\begin{tabular}{lcccccc}
\toprule
\textbf{Model} & \textbf{Parameters} & \textbf{Size} & \textbf{p50 Latency} & \textbf{p90 Latency} & \textbf{Mean Latency} & \textbf{Throughput} \\
& & \textbf{(MB)} & \textbf{(ms)} & \textbf{(ms)} & \textbf{(ms)} & \textbf{(inf./s)} \\
\midrule
TCN-Short & 177,249 & 0.68 & 17.91 & 20.16 & 18.68 & 53.5 \\
TTSNet & 1,657,988 & 6.32 & 11.93 & 12.38 & 12.22 & 81.8 \\
C2L-Net (Proposed) & 161,347 & 0.62 & 0.29 & 0.31 & 0.30 & 3368.4 \\
\bottomrule
\end{tabular}
}
\end{table*}

\subsection{Discussion}
The current experiments are conducted under fixed ambient-temperature conditions, this setup enables a controlled evaluation of the model across distinct thermal regimes. In future work, we will evaluate the
proposed model using real-world data with varying ambient temperatures. We will also extend the model toward a physics-informed neural network to improve prediction accuracy and reduce its dependence on large amounts of training data.

\section{Conclusion}\label{Conclusion}
In this paper, we propose C2L-Net, a Context-to-Latest data-driven model for online SOC estimation of lithium-ion batteries. The proposed model is designed to address the limitations of long-history or heavy data-driven approaches by using only a short historical input window without zero padding at the initial stage, along with a chunking-based architecture. This design reduces the risk of learning artificial positional patterns, enables fast inference, and maintains high SOC estimation accuracy, making it suitable for real-world applications. C2L-Net first divides the short input sequence of current, voltage, and temperature into several chunks to preserve local temporal information while reducing the sequence length processed by the model. The feature extraction module then applies the Theta Attention Pool and Seasonality Basis modules to generate compact tokens for each signal. These tokens are processed by a context encoder consisting of a GRU and Causal Cosine Attention, which captures relevant historical dependencies while preventing future information leakage. Finally, a latest-measurement decoder combines the encoded contextual state with the latest measurement point to generate the SOC estimate. The proposed model is evaluated on a public lithium-ion battery drive-cycle dataset under separate fixed-temperature conditions. The experimental results demonstrate that C2L-Net can accurately estimate SOC using only recent measurements and a lightweight architecture, making it suitable for online applications where long historical profiles are unavailable or computational resources are limited. In future work, we will extend C2L-Net to dynamic-temperature drive profiles to further improve its robustness in more realistic operating environments and incorporate a physics-informed neural network (PINN) framework for enhanced SOC estimation.

\section*{Contact Information}
For access to the code and further information about this proposed system, please contact AIWARE Limited Company at: \url{https://aiware.website/Contact}

\section*{Preprint Availability}
A preprint of this manuscript is available at:
\url{https://arxiv.org/pdf/2605.08653}

% \section*{Conflict of Interest}
% The authors declare that they have no known competing financial interests or personal relationships that could have appeared to influence the work reported in this paper.

% \section*{Data Availability}
% The dataset used during this study is publicly available at: https://doi.org/10.1016/j.est.2024.114888.

\bibliography{sn-bibliography}% common bib file

@article{li2025lightweight,
  title={Lightweight state of charge estimation of lithium-ion battery based on pruned neural networks},
  author={Li, Chaoran and Li, Lele and Zhang, Qiang and Zhou, Shoubin and Li, Menghan and Rao, Zhonghao},
  journal={IEEE Transactions on Transportation Electrification},
  year={2025},
  publisher={IEEE}
}

@article{hu2024state,
  title={State of charge estimation for lithium-ion batteries based on data augmentation with generative adversarial network},
  author={Hu, Chunsheng and Cheng, Fangjuan and Zhao, Yong and Guo, Shanshan and Ma, Liang},
  journal={Journal of Energy Storage},
  volume={80},
  pages={110004},
  year={2024},
  publisher={Elsevier}
}

@article{ge2024structural,
  title={A structural pruning method for lithium-ion batteries remaining useful life prediction model with multi-head attention mechanism},
  author={Ge, Yang and Ma, Jiaxin and Sun, Guodong},
  journal={Journal of Energy Storage},
  volume={86},
  pages={111396},
  year={2024},
  publisher={Elsevier}
}

@article{yang2025physics,
  title={Physics-informed neural network for co-estimation of state of health, remaining useful life, and short-term degradation path in Lithium-ion batteries},
  author={Yang, Li and He, Mingjian and Ren, Yatao and Gao, Baohai and Qi, Hong},
  journal={Applied Energy},
  volume={398},
  pages={126427},
  year={2025},
  publisher={Elsevier}
}

@article{saranathan2025navigating,
  title={Navigating the spectrum of battery health management in electric vehicles: A comprehensive review},
  author={Saranathan, L and Vairavasundaram, Indragandhi and Ashok, Bragadeshwaran},
  journal={Results in Engineering},
  volume={27},
  pages={106038},
  year={2025},
  publisher={Elsevier}
}

@article{cai2024deep,
  title={A deep learning framework for the joint prediction of the SOH and RUL of lithium-ion batteries based on bimodal images},
  author={Cai, Nian and Que, Xiaoping and Zhang, Xu and Feng, Weiguo and Zhou, Yinghong},
  journal={Energy},
  volume={302},
  pages={131700},
  year={2024},
  publisher={Elsevier}
}

@article{wang2025integrated,
  title={Integrated approaches for lithium-ion battery state estimation and life prediction: A critical review of model-driven, data-driven, and hybrid techniques},
  author={Wang, Kunyu and Lin, Xin and Zhang, Xiaodong and Zheng, Jianming and He, Hongzhou and Xu, Ying and Wang, Dechao and Zheng, Zhifeng and Huang, Yuanbo},
  journal={Journal of Cleaner Production},
  volume={521},
  pages={146229},
  year={2025},
  publisher={Elsevier}
}

@article{sun2024state,
  title={State of health estimation for lithium-ion batteries based on current interrupt method and genetic algorithm optimized back propagation neural network},
  author={Sun, Jinghua and Kainz, Josef},
  journal={Journal of Power Sources},
  volume={591},
  pages={233842},
  year={2024},
  publisher={Elsevier}
}

@article{pop2005state,
  title={State-of-the-art of battery state-of-charge determination},
  author={Pop, Valer and Bergveld, Henk Jan and Notten, PHL and Regtien, Paul PL},
  journal={Measurement science and technology},
  volume={16},
  number={12},
  pages={R93--R110},
  year={2005}
}

@article{westerhoff2016electrochemical,
  title={Electrochemical impedance spectroscopy based estimation of the state of charge of lithium-ion batteries},
  author={Westerhoff, Uwe and Kroker, Thorsten and Kurbach, Kerstin and Kurrat, Michael},
  journal={Journal of Energy Storage},
  volume={8},
  pages={244--256},
  year={2016},
  publisher={Elsevier}
}

@article{bage2025enhanced,
  title={Enhanced moving-step unscented transformed-dual extended Kalman filter for accurate SOC estimation of lithium-ion batteries considering temperature uncertainties},
  author={Bage, Alhamdu Nuhu and Takyi-Aninakwa, Paul and Yang, Xiaoyong and Tu, Qingsong Howard},
  journal={Journal of Energy Storage},
  volume={110},
  pages={115340},
  year={2025},
  publisher={Elsevier}
}

@article{he2022comparative,
  title={A comparative study of SOC estimation based on equivalent circuit models},
  author={He, Jiangtao and Meng, Shujuan and Yan, Fengjun},
  journal={Frontiers in Energy Research},
  volume={10},
  pages={914291},
  year={2022},
  publisher={Frontiers Media SA}
}

@article{chen2019particle,
  title={Particle filter-based state-of-charge estimation and remaining-dischargeable-time prediction method for lithium-ion batteries},
  author={Chen, Zonghai and Sun, Han and Dong, Guangzhong and Wei, Jingwen and Wu, JI},
  journal={Journal of Power Sources},
  volume={414},
  pages={158--166},
  year={2019},
  publisher={Elsevier}
}

@article{cui2022extended,
  title={An extended Kalman filter based SOC estimation method for Li-ion battery},
  author={Cui, Zhenjie and Hu, Weihao and Zhang, Guozhou and Zhang, Zhenyuan and Chen, Zhe},
  journal={Energy Reports},
  volume={8},
  pages={81--87},
  year={2022},
  publisher={Elsevier}
}

@article{tang2017observer,
  title={Observer based battery SOC estimation: Using multi-gain-switching approach},
  author={Tang, Xiaopeng and Liu, Boyang and Lv, Zhou and Gao, Furong},
  journal={Applied energy},
  volume={204},
  pages={1275--1283},
  year={2017},
  publisher={Elsevier}
}

@article{chen2025design,
  title={Design of a novel robust UIO estimator with predefined convergence time for the state of charge estimation for lithium-ion batteries},
  author={Chen, Xinzhi and Wang, Kun},
  journal={International Journal of Dynamics and Control},
  volume={13},
  number={1},
  pages={21},
  year={2025},
  publisher={Springer}
}

@article{yao2025multi,
  title={A multi-scale data-driven framework for online state of charge estimation of lithium-ion batteries with a novel public drive cycle dataset},
  author={Yao, Jiaqi and Kowal, Julia},
  journal={Journal of Energy Storage},
  volume={107},
  pages={114888},
  year={2025},
  publisher={Elsevier}
}

@article{oreshkin2019n,
  title={N-BEATS: Neural basis expansion analysis for interpretable time series forecasting},
  author={Oreshkin, Boris N and Carpov, Dmitri and Chapados, Nicolas and Bengio, Yoshua},
  journal={arXiv preprint arXiv:1905.10437},
  year={2019}
}

@article{meng2023machine,
  title={A machine learning approach for face mask detection system with AdamW optimizer},
  author={Meng, Leong Kah and Yi, Ho Hooi and Wei, Ng Bo and Xin, Lim Jia and Salam, Zailan Arabee Abdul and others},
  journal={Journal of Applied Technology and Innovation},
  volume={7},
  number={3},
  year={2023}
}

@article{bao2024ttsnet,
  title={TTSNet: State-of-charge estimation of Li-ion battery in electrical vehicles with temporal transformer-based sequence network},
  author={Bao, Zhengyi and Nie, Jiahao and Lin, Huipin and Gao, Kejie and He, Zhiwei and Gao, Mingyu},
  journal={IEEE Transactions on Vehicular Technology},
  volume={73},
  number={6},
  pages={7838--7851},
  year={2024},
  publisher={IEEE}
}

@article{welch1995introduction,
  title={An introduction to the Kalman filter},
  author={Welch, Greg and Bishop, Gary and others},
  year={1995},
  publisher={Chapel Hill, NC, USA}
}

@inproceedings{mongaras2025cottention,
  title={Cottention: Linear transformers with cosine attention},
  author={Mongaras, Gabriel and Dohm, Trevor and Larson, Eric},
  booktitle={Intelligent Computing-Proceedings of the Computing Conference},
  pages={485--500},
  year={2025},
  organization={Springer}
}

@inproceedings{lea2016temporal,
  title={Temporal convolutional networks: A unified approach to action segmentation},
  author={Lea, Colin and Vidal, Rene and Reiter, Austin and Hager, Gregory D},
  booktitle={European conference on computer vision},
  pages={47--54},
  year={2016},
  organization={Springer}
}

@article{vaswani2017attention,
  title={Attention is all you need},
  author={Vaswani, Ashish and Shazeer, Noam and Parmar, Niki and Uszkoreit, Jakob and Jones, Llion and Gomez, Aidan N and Kaiser, {\L}ukasz and Polosukhin, Illia},
  journal={Advances in neural information processing systems},
  volume={30},
  year={2017}
}

@article{graves2012long,
  title={Long short-term memory},
  author={Graves, Alex},
  journal={Supervised sequence labelling with recurrent neural networks},
  pages={37--45},
  year={2012},
  publisher={Springer}
}

@article{subashini2026physics,
  title={Physics-guided time-series transformer for accurate and robust battery SOC estimation},
  author={Subashini, B and Evangeline, S Ida},
  journal={Neural Computing and Applications},
  volume={38},
  number={8},
  pages={263},
  year={2026},
  publisher={Springer}
}

@inproceedings{dey2017gate,
  title={Gate-variants of gated recurrent unit (GRU) neural networks},
  author={Dey, Rahul and Salem, Fathi M},
  booktitle={2017 IEEE 60th international midwest symposium on circuits and systems (MWSCAS)},
  pages={1597--1600},
  year={2017},
  organization={IEEE}
}

@article{purwono2022understanding,
  title={Understanding of convolutional neural network (cnn): A review},
  author={Purwono, Purwono and Ma'arif, Alfian and Rahmaniar, Wahyu and Fathurrahman, Haris Imam Karim and Frisky, Aufaclav Zatu Kusuma and ul Haq, Qazi Mazhar},
  journal={International Journal of Robotics and Control Systems},
  volume={2},
  number={4},
  pages={739--748},
  year={2022}
}

@article{wong2024balancing,
  title={Balancing accuracy and efficiency: a homogeneous ensemble approach for lithium-ion battery state of charge estimation in electric vehicles},
  author={Wong, Rae Hann and Sooriamoorthy, Denesh and Manoharan, Aaruththiran and Binti Sariff, Nohaidda and Hilmi Ismail, Zool},
  journal={Neural Computing and Applications},
  volume={36},
  number={30},
  pages={19157--19171},
  year={2024},
  publisher={Springer}
}

\end{document}